\documentclass{amia}
\usepackage{lipsum} 

\usepackage{inconsolata}

\usepackage{microtype}

\usepackage{multirow}
\usepackage{multicol}
\usepackage{booktabs}
\usepackage{graphicx}
\usepackage{caption}
\usepackage{subcaption}
\usepackage[inline]{enumitem}
\usepackage{amsmath}
\usepackage[superscript,nomove]{cite}

\usepackage{subcaption}
\usepackage[table]{xcolor}

\usepackage{hyperref}

\DeclareCaptionFormat{boldnumber}{\textbf{#1#2}#3}
\begin{document}

\title{Explainable Diagnosis Prediction through Neuro-Symbolic Integration}

\author{Qiuhao Lu, Ph.D.$^1$, Rui Li, Ph.D.$^1$, Elham Sagheb, M.S.$^2$, Andrew Wen, M.S.$^1$, Jinlian Wang, Ph.D.$^1$, Liwei Wang, M.D., Ph.D.$^1$, Jungwei W. Fan, Ph.D.$^2$, Hongfang Liu, Ph.D.$^1$}

\institutes{$^1$McWilliams School of Biomedical Informatics, The University of Texas Health Science Center, Houston, TX, USA\\$^2$Department of Artificial Intelligence and Informatics, Mayo Clinic, Rochester, MN, USA
}

\maketitle

\section*{Abstract}
\textit{Diagnosis prediction is a critical task in healthcare, where timely and accurate identification of medical conditions can significantly impact patient outcomes. Traditional machine learning and deep learning models have achieved notable success in this domain but often lack interpretability which is a crucial requirement in clinical settings. In this study, we explore the use of neuro-symbolic methods, specifically Logical Neural Networks (LNNs), to develop explainable models for diagnosis prediction. Essentially, we design and implement LNN-based models that integrate domain-specific knowledge through logical rules with learnable weights and thresholds. Our models, particularly $M_{\text{multi-pathway}}$ and $M_{\text{comprehensive}}$, demonstrate superior performance over traditional models such as Logistic Regression, SVM, and Random Forest, achieving higher accuracy (up to 80.52\%) and AUROC scores (up to 0.8457) in the case study of diabetes prediction. The learned weights and thresholds within the LNN models provide direct insights into feature contributions, enhancing interpretability without compromising predictive power. These findings highlight the potential of neuro-symbolic approaches in bridging the gap between accuracy and explainability in healthcare AI applications. By offering transparent and adaptable diagnostic models, our work contributes to the advancement of precision medicine and supports the development of equitable healthcare solutions. Future research will focus on extending these methods to larger and more diverse datasets to further validate their applicability across different medical conditions and populations.}

\section*{Introduction}

Diagnosis prediction is a critical yet challenging task in the healthcare domain, where accurate and timely identification of medical conditions can significantly impact patient outcomes. Over the past few decades, various computational methods have been explored to tackle this problem, evolving from traditional machine learning techniques to more advanced deep learning models and, more recently, large language models (LLMs). Despite their success in specific applications, these methods often lack explainability, which is especially critical in the healthcare domain.

Early efforts in computational diagnosis prediction primarily used symbolic methods, relying on predefined rules and logic to draw conclusions from data \cite{lucas1997symbolic,huang2015interpretable,choubey2017rule}. These methods are highly interpretable, as their reasoning process is transparent. However, their performance is often limited by the rigidity of human-defined rules, which struggle to capture the complexity and variability of real-world medical data. 

Traditional machine learning models, such as decision trees, support vector machines (SVMs), and logistic regression, have improved the ability to analyze medical data by learning patterns from training data \cite{bashir2016hmv,joloudari2019computer,fitriyani2020hdpm,rani2021decision}. These models offer better generalization than symbolic methods and are capable of handling more complex datasets. However, they still require substantial feature engineering, where domain experts must manually select and transform relevant features from raw data. This process is time-consuming and can introduce biases. Moreover, while some models like decision trees are interpretable, others, such as SVMs or ensemble methods (e.g., random forests), can become opaque as the model complexity increases. 

Deep learning models, particularly neural networks, often surpass traditional ML models in terms of accuracy. Convolutional neural networks (CNNs) have been particularly successful in image-based diagnosis, while recurrent neural networks (RNNs) and transformers have shown promise in sequential data, such as electronic health records (EHRs) \cite{ma2017dipole,tufail2021deep,liu2022deep}. However, deep learning models are often criticized for their lack of interpretability or explainability. The ``black-box'' nature has raised concerns about the transparency of decision-making processes, which is especially critical in clinical settings where understanding the rationale behind a diagnosis is crucial. Additionally, deep learning models require large volumes of labeled data for training, which can be challenging to obtain in the healthcare domain.

More recently, large language models (LLMs) have been applied to diagnosis prediction, leveraging their ability to understand and generate human-like text \cite{shoham2023cpllm,gao2023large,koga2024evaluating,warrier2024comparative}. LLMs can process large amounts of textual data, such as clinical notes, and identify complex patterns that might be missed by other models. While LLMs represent a significant advancement in handling unstructured data and can provide contextual insights that are difficult for traditional models to capture, they also inherit many of the challenges associated with deep learning models. LLMs are even more complex and opaque, making it challenging to interpret their predictions. Their massive scale requires substantial computational resources, which can limit their practical application in many healthcare settings. Furthermore, LLMs are not well-suited for working with numerical features or tabular data, which are often crucial in medical contexts, such as lab results or vital signs.

Neuro-symbolic AI (NeSy) is an emerging field that integrates the interpretability and structured reasoning of symbolic methods with the powerful learning capabilities of neural networks. Symbolic reasoning enables the incorporation of domain knowledge and logical rules, providing a level of transparency and explanation that pure neural networks often lack. Meanwhile, neural networks contribute to the model's ability to learn from data and handle complex patterns, making neuro-symbolic AI a compelling candidate for tasks like diagnosis prediction. This hybrid approach is particularly well-suited for healthcare applications where explainability is critical.

In this work, we specifically focus on Logical Neural Networks (LNNs) \cite{riegel2020logical}. LNNs are a neuro-symbolic AI framework designed to learn interpretable models by expressing rules in first-order logic (FOL), a powerful language with clear semantics and a rich set of operators. Unlike traditional neuro-symbolic AI methods that rely on non-learnable $t$-norms \cite{esteva2001monoidal,yang2017differentiable}, LNNs incorporate learnable parameters that allow the network to adjust its logical rules based on data, thereby enhancing its ability to model complex medical information. This makes LNNs not only accurate but also transparent and adaptable, ensuring that the model's decisions are explainable and suitable for clinical settings.

This study represents one of the first explorations into the use of neuro-symbolic methods for diagnosis prediction, a natural fit given the need for both accuracy and explainability. By integrating domain-specific logical rules with neural networks, the approach offers a fresh perspective on the potential of explainable AI in healthcare. The LNN-based models not only outperform traditional machine learning models across multiple metrics but also provide transparent decision-making processes that align with clinical reasoning. In particular, the LNN-based models \( M_{\text{multi-pathway}} \) and \( M_{\text{comprehensive}} \) achieve the highest accuracy of $80.52$\% and AUC scores of $0.8457$ and $0.8399$, respectively, surpassing the best traditional model, Random Forest, which has an accuracy of $76.95$\% and an AUC of $0.8342$. Moreover, \( M_{\text{comprehensive}} \) shows the highest precision of 87.88\%, while \( M_{\text{multi-pathway}} \) achieves the highest F1-score of $68.75$\% among all models tested. This superior performance highlights the effectiveness of integrating logical rules with neural networks, enabling the models to capture complex relationships between risk factors while maintaining interpretability.

In summary, this research contributes to the ongoing discourse on AI in healthcare by demonstrating how neuro-symbolic methods can bridge the gap between performance and explainability. The adaptability of LNNs, with their learned weights and thresholds, creates opportunities for personalized diagnostics and supports the development of precision medicine. While further investigation is necessary to generalize these findings across diverse medical conditions and populations, the insights gained from this study will help build more equitable and transparent AI diagnosis systems in the future.

\section*{Related Work}
Recent years have witnessed a growing interest in developing machine learning methods for diagnosis prediction. For instance, Huang et al. develop a decision-tree-based model that integrates clinical and genetic features with a gender-specific rule to identify the risk of diabetic nephropathy among type 2 diabetes patients \cite{huang2015interpretable}. Ma et al. develop Dipole, a neural method that utilizes bidirectional recurrent neural networks (BiRNNs) to capture and remember information from both past and future patient visits within Electronic Health Records (EHRs) \cite{ma2017dipole}. It introduces three attention mechanisms to model the relationships between different visits, enhancing both prediction accuracy and interpretability of the results. Experimental evaluations on real-world EHR datasets demonstrate that their method outperforms existing diagnosis prediction approaches. Gao et al. introduce Dr.Knows, an innovative approach designed to enhance the capabilities of LLMs in diagnosis prediction by incorporating a medical knowledge graph (KG) derived from the Unified Medical Language System (UMLS) and employing a novel graph model \cite{gao2023large}. This approach leverages the KG to facilitate the interpretation and summarization of complex medical concepts. Experimental results on real-world hospital datasets demonstrate that Dr.Knows not only enhances diagnostic accuracy but also provides an explainable diagnostic pathway. However, despite offering a ``pathway'' for diagnosis prediction, the inherent ``black-box'' nature of neural networks and LLMs poses challenges for fully interpreting and validating these predictions, raising concerns about the trustworthiness of the approach.

In contrast to these methods, which either lack interpretability or provide it only indirectly, we propose a neuro-symbolic approach to diagnosis prediction that offers direct explainability with learnable weights for each diagnostic rule and learnable thresholds for each feature. This approach more effectively meets the critical needs of the healthcare sector by ensuring transparency and interpretability in the diagnostic process.

\begin{table*}[t]
\begin{center}\caption{Statistics of the Pima Indian Diabetes Dataset.}\label{dataset}
\resizebox{0.9\textwidth}{!}{\begin{tabular}{llllll}
    \toprule
    \bf Feature & \bf Description & \bf Data Type & \bf Min & \bf Median & \bf Max\\
    \midrule
    Preg & Number of times pregnant & Numeric & 0.000 & 3.000 & 17.000\\
    Gluc & Plasma glucose concentration (2h GTIT) & Numeric & 0.000 & 117.000 & 199.000\\
    BP & Diastolic Blood Pressure (mm Hg) & Numeric & 0.000 & 72.000 & 122.000\\
    Skin & Triceps skin fold thickness (mm) & Numeric & 0.000 & 23.000 & 99.000\\
    Insulin & 2-Hour Serum insulin (µh/ml) & Numeric & 0.000 & 30.500 & 846.000\\
    BMI & Body mass index (kg/m²) & Numeric & 0.000 & 32.000 & 67.100\\
    DPF & Diabetes pedigree function & Numeric & 0.078 & 0.3725 & 2.4200\\
    Age & Age (years) & Numeric & 21.000 & 29.000 & 81.000\\
    Outcome & Diabetic/Non-diabetic (0 = Non-diabetic, 1 = Diabetic) & Binary & 0 & - & 1\\
    \bottomrule
\end{tabular}}
\end{center}
\end{table*}

\section*{Methods}
\paragraph{Task Overview}
The task of diagnosis prediction involves determining whether a patient has diabetes based on a set of numerical features. Formally, given a dataset of patients, each instance is represented by a feature vector $X = [x_1, x_2, \dots, x_n]$, where $n$ is the number of features, and each feature $x_i$ corresponds to a specific medical measurement (e.g., number of pregnancies, glucose level, BMI). The objective is to predict a binary label $Y \in {0, 1}$, where $Y=1$ indicates the presence of diabetes and $Y=0$ indicates its absence. This task is treated as a binary classification problem, where the model is trained to learn a mapping from the input features $X$ to the target label $Y$.

\paragraph{Dataset}
In our experiments, we use the Pima Indian Diabetes dataset \cite{chang2023pima}, a widely used benchmark dataset drawn from a population with a high incidence of diabetes mellitus. The Pima Indians, a Native American group from Mexico and Arizona, have been the focus of extensive research due to their elevated diabetes prevalence, considered representative of broader global health trends \cite{smith1988using,schulz2006effects}. This dataset is particularly important for global health research and addressing health disparities among underrepresented minority groups. It contains records from $768$ Pima Indian women aged 21 and older, with $258$ testing positive for diabetes and $500$ testing negative. We randomly split the dataset into $80\%$ for training and $20\%$ for testing.

The detailed statistics of the dataset are shown in Table~\ref{dataset}, which provides a summary of the key features used in the dataset. The features include the number of times pregnant (\textbf{Preg}), plasma glucose concentration at 2 hours in an oral glucose tolerance test (\textbf{Gluc}), diastolic blood pressure (\textbf{BP}), triceps skin fold thickness (\textbf{Skin}), 2-hour serum insulin levels (\textbf{Insulin}), body mass index (\textbf{BMI}), diabetes pedigree function (\textbf{DPF}), and age (\textbf{Age}). Additionally, the dataset includes a binary outcome feature (\textbf{Outcome}) indicating whether a person is diabetic (1) or non-diabetic (0). Each feature's data type, along with the minimum, median, and maximum values, are presented to give an overview of the distribution and range of the dataset.


\paragraph{Logical Neural Networks}

Neural networks are powerful tools for learning from data, but they are typically limited in how they handle logic. While classical logic involves operators like conjunction ($\wedge$), disjunction ($\vee$), and negation ($\neg$), these operators are not naturally differentiable. This means that classical Boolean logic is difficult to use with neural networks, which rely on gradient-based methods like backpropagation for learning.

To overcome this challenge, neuro-symbolic AI introduces differentiable versions of logical operators, allowing neural networks to learn from logical rules. For instance, conjunction ($\wedge$) and disjunction ($\vee$) can be replaced with differentiable functions known as $t$-norms and $t$-conorms, respectively \cite{esteva2001monoidal}. One common $t$-norm is the \textit{product} $t$-norm \cite{yang2017differentiable}, which is defined as:
\[
x \wedge y = \text{Product}(x, y) = x \cdot y
\]
where $x, y \in [0,1]$ represent real-valued inputs. This approach works well because it matches classical Boolean logic at the extremes (i.e., when $x, y$ are either $0$ or $1$), but it also allows for smooth changes when $x, y$ take values between $0$ and $1$.

Logical Neural Networks (LNNs) take this idea further by introducing learnable parameters into the logical operators, such as conjunction ($\wedge$), to better adapt to the data \cite{riegel2020logical}. The LNN conjunction operator (LNN-$\wedge$) is defined as:
\[
\max(0, \min(1, \beta - w_1(1-x) - w_2(1-y)))
\]
where $\beta, w_1, w_2$ are learnable parameters, and $x, y \in [0,1]$ are inputs. The output of LNN-$\wedge$ is clamped between $0$ and $1$ to maintain consistency with logical semantics, ensuring that it behaves similarly to classical logic when $x$ and $y$ are close to $0$ or $1$. LNN negation operates as a pass-through function, defined as $\text{LNN-}\neg(x) = 1 - x$. The LNN disjunction operator is derived from the conjunction operator and is defined as:

\[
\text{LNN-}\vee(x, y) = 1 - \text{LNN-}\wedge(1 - x, 1 - y)
\]

Additionally, LNNs impose certain constraints to maintain the crispness of first-order logic (FOL) semantics. These constraints ensure that the learned conjunction behaves properly across all inputs. For example, the constraints in LNN-$\wedge$ are given by:
\[
\beta - (1-\alpha)(w_1 + w_2) \geq \alpha
\]
\[
\beta - \alpha w_1 \leq 1-\alpha
\]
\[
\beta - \alpha w_2 \leq 1-\alpha
\]
where $\alpha \in [\frac{1}{2}, 1]$ is a hyperparameter that controls the strength of the constraint, and $w_1, w_2 \geq 0$. These conditions ensure that the LNN conjunction behaves similarly to classical logic, while still allowing it to learn from data. We direct readers to the foundational work on LNNs for a more comprehensive understanding of their theory \cite{riegel2020logical} and various applications \cite{jiang2021lnn,lu2022cross,sen2022logical}.

\paragraph{LNN-based Diagnosis Prediction}
In this section, we describe the application of LNNs to the task of diagnosis prediction. The task is framed as binary classification that involves predicting whether a patient has a particular medical condition (i.e., diabetes) based on numeric features such as blood pressure, age, gender, etc.

Diagnosis prediction often relies on decision-making rules involving thresholds on patient features. These rules can be expressed in first-order logic (FOL) as a combination of predicates in the form $f_k > \theta_k$, where $f_k$ is a patient feature and $\theta_k$ is a threshold. For example, a simple rule for diagnosing hypertension could be $f_{\text{blood pressure}} > \theta_1 \wedge f_{\text{age}} > \theta_2$. However, such rules are static and require manual tuning.

To enable data-driven learning and flexibility, these logical rules are reformulated into the LNN framework, where conjunctions $(\wedge)$ and disjunctions $(\vee)$ are replaced by their differentiable counterparts: LNN-$\wedge$ and LNN-$\vee$. These operators allow the model to learn from real-valued inputs in $[0,1]$, which are derived from patient features. For each feature, the LNN learns a threshold using a smooth transition logic function $TL(f, \theta) = f \cdot \sigma(f - \theta)$, where $\sigma$ is the sigmoid function and $\theta$ is a learnable threshold ensuring the comparison remains within the $[0,1]$ range \cite{jiang2021lnn}.

For example, the LNN-based hypertension rule becomes:
\[
p(\text{diagnosis=hypertension} \mid \text{patient}) = \text{LNN-}\wedge(TL(f_{\text{blood pressure}}, \theta_1), TL(f_{\text{age}}, \theta_2))
\]
This reformulation ensures the rule is continuous and differentiable, making it suitable for learning via gradient-based optimization. The model can combine multiple rules using LNN-$\vee$ to capture more complex diagnostic scenarios.

The task of diagnosis prediction is framed as a binary classification problem where the model predicts whether a patient has the diagnosis (positive class) or not (negative class). To train the LNN, we minimize the binary cross-entropy loss, defined as:
\[
\mathcal{L}(y, \hat{y}) = - \frac{1}{N} \sum_{i=1}^{N} \left[ y_i \log(\hat{y}_i) + (1 - y_i) \log(1 - \hat{y}_i) \right]
\]
where $y_i$ is the true label for patient $i$, and $\hat{y}_i$ is the predicted probability of the diagnosis.



In general, patient features are passed through the trained LNN model, which applies the learned diagnostic rules to output a probability score for the diagnosis. The LNN's ability to combine learned thresholds and logical rules ensures an interpretable yet flexible approach to diagnosis prediction.


\paragraph{Rule Models for Diabetes Diagnosis Prediction}
For the task of diabetes prediction, we have developed a set of rule-based models that capture key patient features associated with diabetes risk. These rules are expressed using logical operators, and for simplicity, we omit the use of the logical neural network symbol $\text{LNN-}$ and the smooth transition logic function $TL(f, \theta)$:

\[
M_{\text{glucose-bmi}} = f_{\text{gluc}}(x) > \theta_1 \wedge f_{\text{BMI}}(x) > \theta_2
\]
This rule links high glucose levels with obesity, both of which are strong indicators of diabetes.

\[
M_{\text{family-insulin}} = f_{\text{DPF}}(x) > \theta_1 \wedge f_{\text{insulin}}(x) > \theta_2 \wedge f_{\text{age}}(x) > \theta_3
\]
Combining family history, insulin levels, and age provides a more comprehensive assessment of diabetes risk.

\[
M_{\text{balanced}} = \big[ f_{\text{DPF}}(x) > \theta_1 \wedge f_{\text{age}}(x) > \theta_2 \big] \vee \big[ f_{\text{preg}}(x) > \theta_3 \wedge f_{\text{gluc}}(x) > \theta_4 \wedge f_{\text{BP}}(x) > \theta_5 \big] \vee \big[ f_{\text{skin}}(x) > \theta_6 \wedge f_{\text{insulin}}(x) 
\]
\[
> \theta_7 \wedge f_{\text{BMI}}(x) > \theta_8 \big]
\]

This model balances various combinations of risk factors, ensuring that multiple pathways contribute to the risk assessment.

\[
M_{\text{multi-pathway}} = \big[ f_{\text{DPF}}(x) > \theta_1 \wedge f_{\text{age}}(x) > \theta_2 \big] \vee \big[ f_{\text{gluc}}(x) > \theta_3 \wedge f_{\text{insulin}}(x) > \theta_4 \wedge f_{\text{BMI}}(x) > \theta_5 \wedge f_{\text{skin}}(x) > \theta_6 \wedge f_{\text{BP}}(x) 
\]
\[
> \theta_7 \wedge f_{\text{preg}}(x) > \theta_8 \big]
\]
This rule offers a comprehensive assessment, considering both family history and age, or a combination of insulin, glucose, BMI, and other physiological factors.

\[
M_{\text{comprehensive}} = \big[ f_{\text{gluc}}(x) > \theta_1 \wedge f_{\text{insulin}}(x) > \theta_2 \wedge f_{\text{BMI}}(x) > \theta_3 \wedge f_{\text{preg}}(x) > \theta_4 \big] \vee \big[ f_{\text{DPF}}(x) > \theta_5 \wedge f_{\text{insulin}}(x) > \theta_6 \big]
\]
This model operates through two pathways: one assessing metabolic factors and another linking family history with insulin levels. This structure allows the model to capture risk from both immediate physiological conditions and hereditary background.


By structuring these models with logical combinations of clinically significant features, we aim to enhance both the interpretability and predictive power of diabetes diagnosis models. Each model captures different aspects of diabetes risk, from immediate physiological conditions to long-term hereditary factors, providing a diverse and comprehensive approach to risk assessment.

\begin{table*}[t]
\begin{center}
\caption{Performance comparison of models on various metrics.}\label{table:model_performance}
\begin{tabular}{lccccc}
    \toprule
    \bf Model & \bf Accuracy & \bf Precision & \bf Recall & \bf F1 & \bf AUC \\
    \midrule
    Logistic Regression & 0.7617 & 0.7283 & 0.5121 & 0.5980 & 0.8262 \\
    SVM & 0.7669 & 0.7154 & 0.5519 & 0.6207 & 0.8315 \\
    Random Forest & 0.7695 & 0.7072 & 0.5876 & 0.6380 & 0.8342 \\
    KNN & 0.7110 & 0.6017 & 0.5053 & 0.5474 & 0.7659 \\
    Naive Bayes & 0.7539 & 0.6645 & \bf0.6011 & 0.6281 & 0.8140 \\
    \midrule
    $M_{\text{glucose-bmi}}$ & 0.7338 & 0.7692 & 0.3636 & 0.4938 & 0.8035 \\
    $M_{\text{family-insulin}}$ & 0.6494 & 0.6667 & 0.0364 & 0.0690 & 0.6509 \\
    $M_{\text{balanced}}$ & 0.7922 & 0.8108 & 0.5455 & 0.6522 & 0.8257 \\
    $M_{\text{multi-pathway}}$ & \bf0.8052 & 0.8049 & 0.6000 & \bf0.6875 &\bf 0.8457 \\
    $M_{\text{comprehensive}}$ & \bf0.8052 & \bf0.8788 & 0.5273 & 0.6591 & 0.8399 \\
    \bottomrule
\end{tabular}
\end{center}
\end{table*}

\paragraph{Evaluation}

We evaluate a set of baseline models, including Logistic Regression, SVM, Random Forest, K-Nearest Neighbors (KNN), and Naive Bayes, using Accuracy, Precision (P), Recall (R), F1-score (F1), and Area Under the Receiver Operating Characteristic Curve (AUC) as metrics. These traditional models are specifically chosen for their interpretability, as each provides a degree of transparency in its decision-making process, which is critical for understanding diagnosis predictions. Their interpretability makes them well-suited for comparison with the five LNN-based models, which integrate diverse features and rules. This evaluation setup allows for a comprehensive comparison between interpretable traditional methods and neuro-symbolic approaches, promoting a deeper understanding of the decision-making mechanisms underlying diagnosis predictions.

\section*{Results}

Table~\ref{table:model_performance} shows the performance comparison of both traditional models and LNN-based models for diabetes diagnosis prediction.

\paragraph{Traditional Models}
The traditional models—including Logistic Regression, SVM, Random Forest, KNN, and Naive Bayes—exhibit varying degrees of performance. The Random Forest model achieves the highest accuracy ($0.7695$), F1-score ($0.6380$), and AUC ($0.8342$) among the baseline models, indicating a strong balance between precision and recall. This performance can be attributed to Random Forest's ability to handle nonlinear relationships and interactions between features. SVM and Logistic Regression also demonstrate competitive performance, with AUCs of $0.8315$ and $0.8262$, respectively. Naive Bayes, while achieving the highest recall ($0.6011$) among the baselines, has a lower precision ($0.6645$), leading to an F1-score of $0.6281$. KNN shows the lowest performance across all metrics among the baseline models, suggesting its limited ability in this task.

\paragraph{LNN-based Models}
The LNN-based models generally outperform the baseline models across multiple metrics. These models leverage logical combinations of key features related to diabetes risk factors, capturing more complex relationships in the data.

\begin{itemize}
    \item $M_{\text{glucose-bmi}}$: While it achieves moderate accuracy ($0.7338$) and AUC ($0.8035$) among the LNN models, it has a low recall ($0.3636$) and F1-score ($0.4938$). This is expected since the model only considers two features, potentially overlooking other important risk factors.
    
    \item $M_{\text{family-insulin}}$: The model performs poorly across all metrics. The very low recall ($0.0364$) indicates that the model fails to identify most positive cases. This may be due to the strictness of the conjunction of three conditions, which significantly narrows the criteria for predicting diabetes, thereby reducing sensitivity.

    \item $M_{\text{balanced}}$: This model shows improved performance, with an accuracy of $0.7922$ and an F1-score of $0.6522$. The use of disjunctions allows the model to capture patients who meet any one of several sets of criteria, enhancing its ability to identify positive cases, as reflected in its higher recall ($0.5455$) compared to $M_{\text{glucose-bmi}}$ and $M_{\text{family-insulin}}$.

    \item $M_{\text{multi-pathway}}$: This model achieves the highest F1-score ($0.6875$) and AUROC ($0.8457$) among all models, indicating a strong and balanced performance. By integrating both hereditary factors and a comprehensive set of physiological metrics, $M_{\text{multi-pathway}}$ effectively captures a wider range of diabetes risk profiles.

    \item $M_{\text{comprehensive}}$: The model achieves the highest precision ($0.8788$) among all models and shares the highest accuracy ($0.8052$) with $M_{\text{multi-pathway}}$. The high precision indicates that the model is highly effective at correctly identifying patients without false positives. However, its recall ($0.5273$) is slightly lower than that of $M_{\text{multi-pathway}}$, suggesting it may miss some positive cases due to the specificity of its rules.

\end{itemize}

\begin{figure}[htbp]
    \centering
    \begin{subfigure}[b]{0.5\textwidth}
        \centering
        \includegraphics[width=\textwidth]{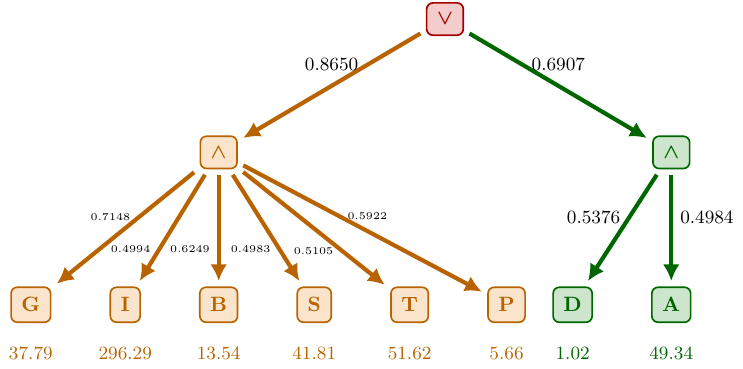}
        \caption{\( M_{\text{multi-pathway}} \)}
        \label{fig:multi-pathway}
    \end{subfigure}
    \hfill 
    \begin{subfigure}[b]{0.48\textwidth}
        \centering
        \includegraphics[width=\textwidth]{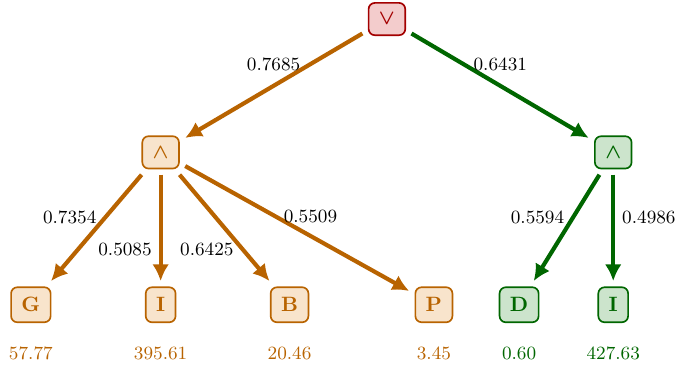}
        \caption{\( M_{\text{comprehensive}} \)}
        \label{fig:comprehensive}
    \end{subfigure}
    \caption{
    Visualization of feature contributions in the rule models \( M_{\text{multi-pathway}} \) and \( M_{\text{comprehensive}} \) for diabetes prediction. The features are denoted as follows: G (glucose), I (insulin), B (BMI), S (skin thickness), T (blood pressure), D (DPF), and A (age). Both diagrams represent the logical structures of the models, where the leaf nodes correspond to features with their respective learned thresholds displayed below each node. The edges connecting the nodes are labeled with weights, indicating the relative importance of each feature in the prediction process. The combination of feature weights and thresholds demonstrates how different pathways contribute to the final prediction in each model.}
    \label{fig:rule-models}
\end{figure}

\section*{Analysis}

The results demonstrate that the LNN-based models, particularly $M_{\text{multi-pathway}}$ and $M_{\text{comprehensive}}$, provide significant improvements over traditional models. Their superior performance can be attributed to the incorporation of domain-specific knowledge through logical rules, which allows them to model complex relationships between features relevant to diabetes.

\begin{itemize}
    \item \textbf{Impact of Model Complexity}: The models that consider multiple pathways (\( M_{\text{balanced}} \), \( M_{\text{multi-pathway}} \), and \( M_{\text{comprehensive}} \)) outperform simpler models (\( M_{\text{glucose-bmi}} \) and \( M_{\text{family-insulin}} \)). This suggests that capturing the heterogeneous nature of diabetes risk factors leads to better predictive performance.
    
    \item \textbf{Role of Logical Operators}: The use of disjunctions ($\vee$) in \( M_{\text{balanced}} \), \( M_{\text{multi-pathway}} \), and \( M_{\text{comprehensive}} \) allows these models to capture patients who meet any of several sets of criteria. This flexibility increases the models' recall by identifying more true positive cases.
    
    \item \textbf{Underperformance of \( M_{\text{family-insulin}} \)}: The poor performance of this model highlights the limitations of relying on a narrow set of features connected strictly by conjunctions ($\wedge$). The strict criteria reduce the model's ability to identify positive cases, leading to low recall.
    
    \item \textbf{Comparative Advantage over Baselines}: The LNN-based models outperform the baseline models not only in the numeric metrics but also in maintaining interpretability and explainability. While Random Forest and SVM provide strong performance among baselines, they lack the transparency of LNN-based models, which is critical in healthcare applications for ethical and practical reasons. 
    
    \item \textbf{Interpretability} Figure~\ref{fig:rule-models} shows the rule models \( M_{\text{multi-pathway}} \) and \( M_{\text{comprehensive}} \) for diabetes prediction. Both models share the same architecture—a logical disjunction (\( \vee \)) of two conjunctions (\( \wedge \))—indicating that diabetes can be predicted through multiple pathways. The key difference lies in the selection of features within each conjunction.

    In \( M_{\text{multi-pathway}} \), the left conjunction includes glucose (G), insulin (I), BMI (B), skin thickness (S), blood pressure (T), and pregnancies (P), all of which are clinical measures associated with metabolic health. The right conjunction incorporates Diabetes Pedigree Function (D) and age (A), highlighting genetic predisposition and demographic factors. In contrast, \( M_{\text{comprehensive}} \) focuses on a subset of these features, emphasizing the combined effect of glucose, insulin, BMI, and pregnancies in the left conjunction, and insulin and DPF in the right.

    Analyzing the weights and thresholds, we observe that features like glucose and BMI have significant weights in both models, aligning with medical knowledge that elevated glucose levels and high BMI are strong indicators of diabetes risk. Notably, the learned thresholds for these features are often below the median values in the dataset (e.g., glucose thresholds below the median of $117$ as mentioned in Table~\ref{dataset}), suggesting that even modest elevations in these features can contribute to a positive prediction, compensating for the strictness of the conjunction. Conversely, in \( M_{\text{comprehensive}} \), the left conjunction includes fewer features, and the thresholds are higher, suggesting more significant deviations from normal values for the condition to be satisfied, which balances the less stringent conjunction (fewer features need to meet their thresholds). This insight illustrates how the models balance feature selection and threshold settings to capture subtle but clinically relevant patterns, enhancing their interpretability and potential utility in early detection of diabetes.
\end{itemize}

In summary, the integration of domain-specific logical rules in the LNN-based models enhances their ability to predict diabetes accurately while maintaining interpretability. The models that consider multiple pathways and balance various risk factors, such as \( M_{\text{multi-pathway}} \) and \( M_{\text{comprehensive}} \), demonstrate the highest effectiveness. These findings underscore the potential of neuro-symbolic approaches in developing predictive models that are both accurate and interpretable, offering valuable tools for clinical decision-making.

\section*{Discussion and Conclusion}



In this study, we explore the integration of neuro-symbolic AI methods, specifically Logical Neural Networks (LNNs), for explainable diagnosis prediction, with a focus on diabetes. The results demonstrate that neuro-symbolic approaches offer a promising balance between accuracy and explainability, addressing the critical need for transparent models in healthcare. By incorporating logical rules based on domain-specific knowledge and learning thresholds from data, the LNN-based models are able to outperform traditional machine learning models, such as logistic regression, SVM, and random forest. Specifically, models like \( M_{\text{multi-pathway}} \) and \( M_{\text{comprehensive}} \) balance multiple pathways of risk factors, capturing complex interdependencies between physiological and hereditary factors associated with diabetes. These models demonstrate high performance while maintaining interpretability, making them natural fits for real-world applications in healthcare.

A significant advantage of our approach lies in the interpretability of the learned weights and thresholds within the LNN-based rule models. These parameters provide direct insight into how different features contribute to the diagnosis prediction. For example, the learned thresholds for features like glucose levels and BMI are consistent with clinical knowledge, highlighting their importance in diabetes risk assessment. This transparency is crucial for clinical settings, as it allows healthcare professionals to understand, trust, and even adjust the model's decision-making process.

Moreover, the adaptability of neuro-symbolic approaches has important implications for healthcare equity. Different patient cohorts may exhibit varying risk factors and thresholds due to genetic, environmental, or socioeconomic differences. By learning these thresholds directly from data, LNNs can tailor diagnostic models to specific populations, enhancing the accuracy and fairness of predictions across diverse groups. This capability supports the development and implementation of precision medicine, where treatments and diagnostics are customized to individual patient characteristics, potentially improving outcomes for underrepresented and underserved populations.

Despite these promising findings, there are limitations to our study. The Pima Indian Diabetes dataset, while widely used, represents a single disease and a relatively small cohort. The generalizability of our neuro-symbolic approach to other diseases or more diverse populations requires further exploration. Moreover, while LNNs have shown their effectiveness in modeling diabetes risk factors, the complexity of certain rules, such as those relying on conjunctions, may limit the sensitivity of the model under specific conditions, as seen with $M_{\text{family-insulin}}$. Additionally, the design of the rules in LNNs relies heavily on domain expertise, which introduces potential biases and limits scalability when applied to other medical conditions where expert knowledge may be less established or more varied.

In future work, we plan to extend the application of neuro-symbolic methods to more comprehensive and larger healthcare datasets, covering a wider range of medical conditions. This would allow us to further assess the scalability and flexibility of neuro-symbolic approaches in diagnosis prediction across diverse clinical contexts. Additionally, we aim to investigate the potential of these methods in enhancing diagnostic accuracy for underrepresented groups, which is crucial for the advancement of precision medicine.

In conclusion, this study highlights the potential of neuro-symbolic methods, particularly Logical Neural Networks, in advancing explainable and accurate diagnosis prediction models. By combining the strengths of symbolic reasoning and neural networks, LNNs offer a viable solution to the challenges of interpretability and performance in healthcare AI applications. The learned weights and thresholds not only improve model transparency but also provide opportunities to address healthcare disparities through personalized diagnostics. As we continue to explore and refine these methods, neuro-symbolic AI holds promise for contributing to more equitable, reliable, and effective healthcare solutions in the future.

\subparagraph{Acknowledgments}
This work was conducted under support from the National Human Genome Research Institute (R01HG12748), the National Library of Medicine (R01LM011934) and the Cancer Prevention and Research Institute of Texas (RR230020).

\makeatletter
\renewcommand{\@biblabel}[1]{\hfill #1.}
\makeatother

\bibliographystyle{vancouver}
\bibliography{amia}  

\begin{thebibliography}{10}

\bibitem{lucas1997symbolic}
Lucas P.
\newblock Symbolic diagnosis and its formalisation.
\newblock The Knowledge Engineering Review. 1997;12(2):109-46.

\bibitem{huang2015interpretable}
Huang GM, Huang KY, Lee TY, Weng JTY.
\newblock An interpretable rule-based diagnostic classification of diabetic nephropathy among type 2 diabetes patients.
\newblock In: BMC bioinformatics. vol.~16. Springer; 2015. p. 1-10.

\bibitem{choubey2017rule}
Choubey DK, Paul S, Dhandhenia VK.
\newblock Rule based diagnosis system for diabetes.
\newblock An International Journal of Medical Sciences. 2017;28(12):5196-208.

\bibitem{bashir2016hmv}
Bashir S, Qamar U, Khan FH, Naseem L.
\newblock HMV: A medical decision support framework using multi-layer classifiers for disease prediction.
\newblock Journal of Computational Science. 2016;13:10-25.

\bibitem{joloudari2019computer}
Joloudari JH, Saadatfar H, Dehzangi A, Shamshirband S.
\newblock Computer-aided decision-making for predicting liver disease using PSO-based optimized SVM with feature selection.
\newblock Informatics in medicine unlocked. 2019;17:100255.

\bibitem{fitriyani2020hdpm}
Fitriyani NL, Syafrudin M, Alfian G, Rhee J.
\newblock HDPM: an effective heart disease prediction model for a clinical decision support system.
\newblock IEEE Access. 2020;8:133034-50.

\bibitem{rani2021decision}
Rani P, Kumar R, Ahmed NMS, Jain A.
\newblock A decision support system for heart disease prediction based upon machine learning.
\newblock Journal of Reliable Intelligent Environments. 2021;7(3):263-75.

\bibitem{ma2017dipole}
Ma F, Chitta R, Zhou J, You Q, Sun T, Gao J.
\newblock Dipole: Diagnosis prediction in healthcare via attention-based bidirectional recurrent neural networks.
\newblock In: Proceedings of the 23rd ACM SIGKDD international conference on knowledge discovery and data mining; 2017. p. 1903-11.

\bibitem{tufail2021deep}
Tufail AB, Ma YK, Kaabar MK, Mart{\'\i}nez F, Junejo A, Ullah I, et~al.
\newblock Deep learning in cancer diagnosis and prognosis prediction: a minireview on challenges, recent trends, and future directions.
\newblock Computational and Mathematical Methods in Medicine. 2021;2021(1):9025470.

\bibitem{liu2022deep}
Liu T, Siegel E, Shen D.
\newblock Deep learning and medical image analysis for COVID-19 diagnosis and prediction.
\newblock Annual review of biomedical engineering. 2022;24(1):179-201.

\bibitem{shoham2023cpllm}
Shoham OB, Rappoport N.
\newblock Cpllm: Clinical prediction with large language models.
\newblock arXiv preprint arXiv:230911295. 2023.

\bibitem{gao2023large}
Gao Y, Li R, Croxford E, Tesch S, To D, Caskey J, et~al.
\newblock Large language models and medical knowledge grounding for diagnosis prediction.
\newblock medRxiv. 2023:2023-11.

\bibitem{koga2024evaluating}
Koga S, Martin NB, Dickson DW.
\newblock Evaluating the performance of large language models: ChatGPT and Google Bard in generating differential diagnoses in clinicopathological conferences of neurodegenerative disorders.
\newblock Brain Pathology. 2024;34(3):e13207.

\bibitem{warrier2024comparative}
Warrier A, Singh R, Haleem A, Zaki H, Eloy JA.
\newblock The comparative diagnostic capability of large language models in otolaryngology.
\newblock The Laryngoscope. 2024.

\bibitem{riegel2020logical}
Riegel R, Gray A, Luus F, Khan N, Makondo N, Akhalwaya IY, et~al.
\newblock Logical neural networks.
\newblock arXiv preprint arXiv:200613155. 2020.

\bibitem{esteva2001monoidal}
Esteva F, Godo L.
\newblock Monoidal t-norm based logic: towards a logic for left-continuous t-norms.
\newblock Fuzzy sets and systems. 2001;124(3):271-88.

\bibitem{yang2017differentiable}
Yang F, Yang Z, Cohen WW.
\newblock Differentiable learning of logical rules for knowledge base reasoning.
\newblock Advances in neural information processing systems. 2017;30.

\bibitem{chang2023pima}
Chang V, Bailey J, Xu QA, Sun Z.
\newblock Pima Indians diabetes mellitus classification based on machine learning (ML) algorithms.
\newblock Neural Computing and Applications. 2023;35(22):16157-73.

\bibitem{smith1988using}
Smith JW, Everhart JE, Dickson W, Knowler WC, Johannes RS.
\newblock Using the ADAP learning algorithm to forecast the onset of diabetes mellitus.
\newblock In: Proceedings of the annual symposium on computer application in medical care. American Medical Informatics Association; 1988. p. 261.

\bibitem{schulz2006effects}
Schulz LO, Bennett PH, Ravussin E, Kidd JR, Kidd KK, Esparza J, et~al.
\newblock Effects of traditional and western environments on prevalence of type 2 diabetes in Pima Indians in Mexico and the US.
\newblock Diabetes care. 2006;29(8):1866-71.

\bibitem{jiang2021lnn}
Jiang H, Gurajada S, Lu Q, Neelam S, Popa L, Sen P, et~al.
\newblock LNN-EL: A Neuro-Symbolic Approach to Short-text Entity Linking.
\newblock In: Proceedings of the 59th Annual Meeting of the Association for Computational Linguistics and the 11th International Joint Conference on Natural Language Processing (Volume 1: Long Papers); 2021. p. 775-87.

\bibitem{lu2022cross}
Lu Q, Gurajada S, Sen P, Popa L, Dou D, Nguyen T.
\newblock Cross-lingual short-text entity linking: Generating features for neuro-symbolic methods.
\newblock In: Proceedings of the fourth workshop on data science with human-in-the-loop (language advances); 2022. p. 8-14.

\bibitem{sen2022logical}
Sen P, Carvalho BW, Abdelaziz I, Kapanipathi P, Roukos S, Gray A.
\newblock Logical neural networks for knowledge base completion with embeddings \& rules.
\newblock In: Proceedings of the 2022 Conference on Empirical Methods in Natural Language Processing; 2022. p. 3863-75.

\end{thebibliography}

\end{document}